\documentclass[runningheads]{llncs}

\usepackage{graphicx}
\usepackage{comment}
\usepackage{amsmath,amssymb}
\usepackage{color}
\usepackage{url}
\usepackage{hyperref}
\usepackage{xspace}

\providecommand{\myparagraph}[1]{{\noindent\textbf{#1}}}

\providecommand{\newsplitname}{TruZe\xspace}
\newif\ifreview
\reviewfalse

\ifreview
	\usepackage{lineno}

	\linenumbers
\fi

\begin{document}


\def\SubNumber{69}

\def\GCPRTrack{Regular Track}

\title{A New Split for Evaluating \\ True Zero-Shot Action Recognition} 

\ifreview
\else

	\author{Shreyank N Gowda\inst{1} \and
	Laura Sevilla-Lara\inst{1} \and
	Kiyoon Kim\inst{1} \and
	Frank Keller\inst{1} \and
	Marcus Rohrbach\inst{2}}
	
	\authorrunning{S. N. Gowda et al.}
	
	\institute{University of Edinburgh, UK \and Facebook AI Research, US
	}
\fi

\maketitle              
\begin{abstract}
Zero-shot action recognition is the task of classifying action categories that are not available in the training set. 
In this setting, the standard evaluation protocol is to use existing action recognition datasets (e.g. UCF101) and \emph{randomly} split the classes into seen and unseen. However, most recent work builds on representations pre-trained on the Kinetics dataset, where classes largely overlap with classes in the zero-shot evaluation datasets. As a result, classes which are supposed to be unseen, are present during supervised pre-training, invalidating the condition of the zero-shot setting.
A similar concern was previously noted several years ago for image based zero-shot recognition, but has 
not been considered by the zero-shot \emph{action} recognition community.  
In this paper, we propose a new split for \emph{true} zero-shot action recognition with 
no overlap between unseen test classes and training or pre-training classes. 
We benchmark several recent approaches on the proposed True Zero-Shot (\textbf{\newsplitname}) Split  for UCF101 and HMDB51, with zero-shot and generalized zero-shot evaluation. In our extensive analysis we find that our \newsplitname{} splits are significantly harder than comparable random splits as nothing is leaking from pre-training, i.e. unseen performance is consistently lower, up to 8.9\% for zero-shot action recognition. In an additional evaluation we also find that similar issues exist in the splits used in few-shot action recognition, here we see differences of up to 17.1\%. We publish our splits \footnote[1]{Splits can be found at \url{https://github.com/kini5gowda/TruZe}} and hope that our benchmark analysis will change how the field is evaluating zero- and few-shot action recognition moving forward.

\end{abstract}

\section{Introduction}

Much of the recent progress in action recognition is due to 
the availability of large annotated datasets. Given how impractical it is to obtain thousands of videos in order to recognize a single class label, researchers have turned to the problem of zero-shot learning (ZSL). Each class label has semantic embeddings that are either manually annotated or inferred through semantic knowledge using word embeddings. These embeddings help obtain relationships between training classes (that have many samples) and test classes (that have zero samples). Typically, the model predicts the semantic embedding of the input video and matches it to a test class using the nearest neighbor's search.


However, work in video ZSL~\cite{e2e,od,claster} often uses a pre-trained model to represent videos. 
While pre-trained models help obtaining good visual representations, 
overlap with test classes can invalidate the premise of zero-shot learning, making it difficult to compare approaches fairly. 

\begin{figure}[t]
    \centering
    \includegraphics[width=1.0\linewidth]{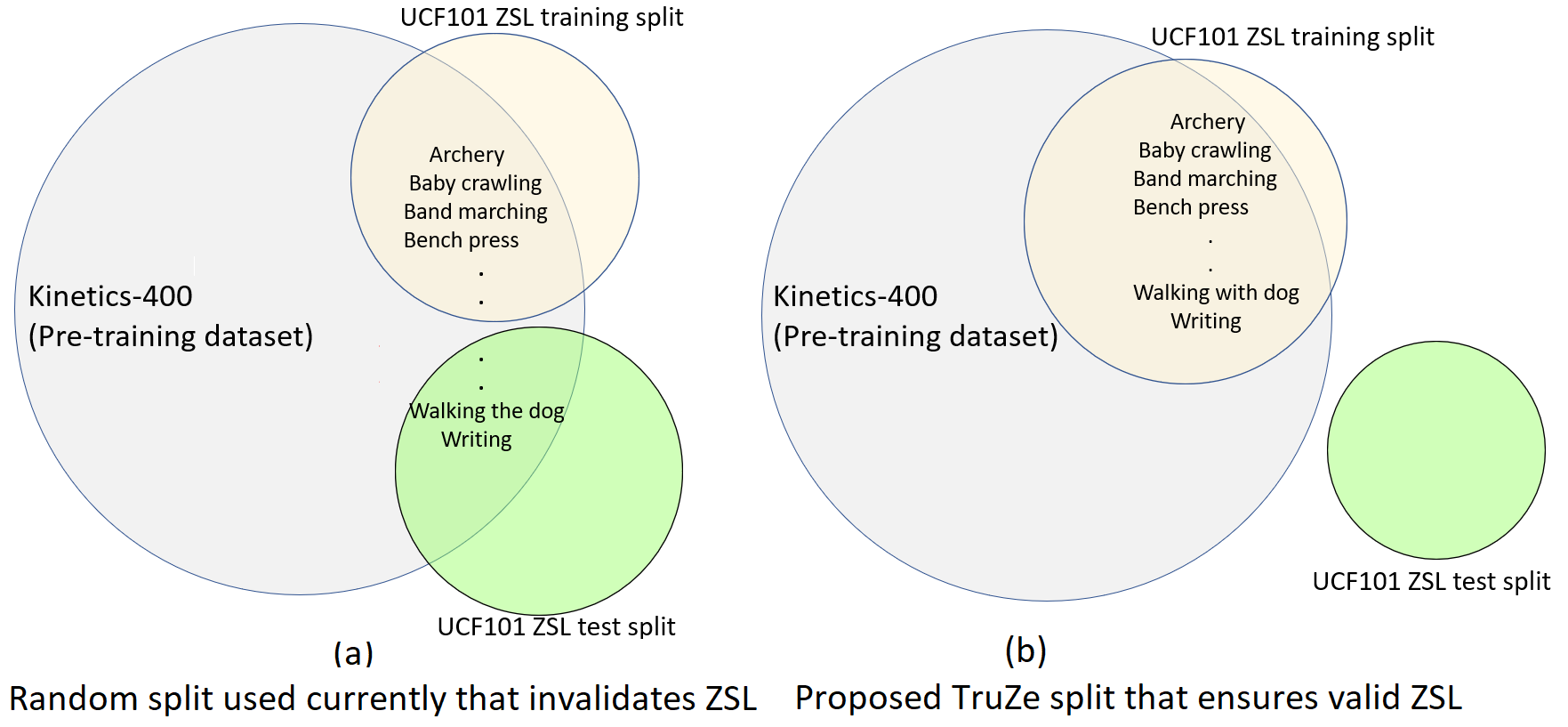}
    \caption{An illustration of the overlap in \emph{classes} of the pretraining dataset (grey), training split (yellow) and zero-shot test split (green). 
    Current evaluation protocol (a) picks classes at random with 51 training classes and 50 test classes. There is always some overlap and also chances of an extremely high overlap. We propose a stricter evaluation protocol (b) where there is no overlap at test time, maintaining the ZSL premise.}
    \label{fig:overlap}
\end{figure}

In the image domain  ~\cite{zsli1,zsli2,bidilel,gbu}, this problem has also been observed. Typically image models are pre-trained on ImageNet \cite{imagenet}. Xian et al.~\cite{gbu} showed that, in image ZSL, if the pre-training dataset has overlapping classes with the test set, the accuracy is inflated at test time. 
Hence, the authors propose a new split that avoids that problem, and it is now widely used. Similarly,  most video models are pre-trained on Kinetics-400 \cite{i3d}, which has a large overlap with the typical ZSL action recognition benchmarks (UCF101, HMDB51 and Olympics). This pre-training gives leads to inflated accuracies, creating the need for a new split. Figure~\ref{fig:overlap} shows an illustration of these overlap issues.

\textbf{Contributions}: First, we show the significant difference in performance caused by pre-training on classes that are included in the test set, across all networks and all datasets. Second, we measure the extent of the overlap between Kinetics-400 and the datasets typically used for ZSL testing: UCF101, HMDB51 and Olympics datasets. We do this by computing both visual and semantic similarity between classes. Finally, we propose a fair split of the classes that takes this class overlap into account, and does not break the premise of ZSL. We show that current models do indeed perform more poorly in this split, which is further proof of the significance of the problem. We hope that this split will be useful to the community, will avoid the need of random splits, and help an actually fair comparison among methods.



\section{Related Work}

Previous work~\cite{gbu} has studied the effect of pre-training on classes that overlap with test classes in the image domain. The authors compute the extent of overlap between testing datasets and Imagenet \cite{imagenet}, where models are typically pre-trained. The overlapping classes correspond to the training classes, while the non-overlapping classes correspond to the test classes. Figure~\ref{fig:overlap} shows an illustration of the proposed evaluation protocol. Unlike the traditional evaluation protocol that chooses classes at random from the list of classes, without typically looking at the list of overlapping classes from the pre-trained dataset, we strictly remove all classes that have a high threshold of visual or semantic similarity (see Sec.~\ref{sec:eval}).

Roitberg et al. \cite{roitberg2018towards} proposed to look at the overlapping classes in videos by using a corrective method that would automatically remove categories that are similar. This was done by utilizing a pairwise similarity of labels within the same dataset. While they showed that using pre-trained models resulted in improved accuracy due to class overlap, the evaluation included only one dataset, and only looked at the semantic similarity of labels. Adding visual similarity, helps discovering overlapping classes like ``typing" in UCF101 and ``using computer" in Kinetics. 
Therefore, in our proposed split we use both semantic and visual similarity across classes. 

Busto et al. \cite{busto2018open} provide a mapping of shared classes between UCF101 and Kinetics as part of a domain adaptation problem. They manually find semantic matches based on class names. However, their mapping was not based on visual and semantic similarity as they have classes such as typing and writing on board as part of UCF101 classes not similar to any Kinetics class. Also, they use "floor gymnastics" as an unknown class in UCF101, however, Kinetics has "gymnastics tumbling" which is the same action. We see that samples from the "typing" class consist of a large proportion of people using their computers and this maps directly to the "using computers" class in Kinetics. Based on our visual and semantic similarity approach, we obtain a slightly different set of classes to those proposed by Busto et al.

Recently, end-to-end training \cite{e2e} has been proposed for ZSL in video classification. As part of the evaluation protocol, to uphold the ZSL premise, the authors propose to train a model on Kinetics by removing the set of overlapping classes (using semantic matching) and using this as a pre-trained model. While this is a promising way to ensure the following of the premise of ZSL, it is very computationally expensive. We also show that having a better backbone (see Sec~\ref{sec:backbone}) results in better accuracy, and as such, training end-to-end is expensive. As a result, using a proposed split instead whilst having the opportunity to use any backbone seems an easier approach.

\section{ZSL preliminaries}

Consider $S$ to be the training set of seen classes, composed of tuples $\left( x, y, a(y) \right)$, where $x$ represents the visual features of each sample in $S$ (spatio-temporal features in the case of video),
$y$ corresponds to the class label in the set of seen class labels $Y_{s}$, and $a(y)$ represents the semantic embedding of class $y$. These semantic embeddings are either annotated manually or computed using a language-based embedding, e.g. word2vec \cite{w2v} or sen2vec \cite{s2v}. 

Let $U$ be the set of unseen classes, composed of tuples $ (u, a(u))$, where $u$ is a class in the label set $Y_{u}$, and $a(u)$ are the corresponding semantic representations. 
$Y_s$ and $Y_u$ do not overlap, i.e.
\begin{equation}
    \label{eq:zeroshot}
    Y_s\bigcap Y_u = \emptyset
\end{equation}

In ZSL for video classification, given an input video, the task is to predict a class label in the unseen set of classes, $f_{ZSL}:X\rightarrow Y_{u}$. An extension of the problem is the related generalized zero-shot learning (GZSL) setting, where given a video, the task is to predict a class label in the union of the seen and unseen classes, as $f_{GZSL}:X\rightarrow Y_{s}\cup Y_{u}$. 

When relying on a pre-trained model to obtain visual features, we denote the pre-trained classes as the set $Y_p$. For the ZSL premise to be maintained, there must be no overlap with the unseen classes: 
\begin{equation}
    \label{eq:truezeroshot}
    Y_p \bigcap Y_u = \emptyset.
\end{equation}

The core problem we address in this paper is that while prior work generally adheres to Eq.~\ref{eq:zeroshot}, recent use of pre-trained models does not adhere to Eq.~\ref{eq:truezeroshot}.  Instead, we propose the \newsplitname split in Section \ref{sec:truezeroshotsplit}, which adheres to both Eq.~(\ref{eq:zeroshot}) \emph{and} Eq.~(\ref{eq:truezeroshot}).


\subsection{Visual and Semantic Embeddings}

Early work computed {\bf visual embeddings} (or representations) using hand-crafted features such as Improved Dense Trajectories (IDT)~\cite{idt}, which include tracked trajectories of detected interest points within a video, and four descriptors. More recent work often uses deep features such as those from 3D convolutional networks (e.g., I3D \cite{i3d} or C3D \cite{c3d}). These 3D CNNs are used to learn spatio-temporal representation of the video. In our experiments, we will use both types of visual representations. 


To obtain {\bf semantic embeddings}, previous work \cite{attr} uses manual attribute annotations for each class. For example, the action of kicking would have motion of the leg and motion of twisting the upper body. However, such attributes are not available for all datasets. An alternative approach is to use word embeddings such as word2vec \cite{w2v} for each class label. This gets rid of the requirement of manual attributes. More recently, Gowda et al.\cite{claster} showed that using sen2vec \cite{s2v} instead of word2vec yields better results as action labels are typically multi-worded and averaging them using word2vec makes it lose context. Based on this, in our experiments we use sen2vec.

\section{Evaluated Methods}

We consider early approaches that use IDT features such as ZSL by bi-directional latent embedding learning (BiDiLEL), ZSL by single latent embedding (Latem) \cite{latem} and synthesized classifiers (SYNC) \cite{sync}. Using features that are not learned, allows us to control for the effect of pre-training when using random splits, and when using the proposed split (PS). 
We then evaluate recent state-of-the-art approaches such as feature generating networks (WGAN) \cite{wgan}, out-of-distribution detection networks (OD) \cite{od} and end-to-end learning for ZSL (E2E) \cite{e2e} as well. Let us briefly have a look at these methods.

\myparagraph{Latem \cite{latem}} uses piece-wise
linear compatibility to understand the visual-semantic embedding relationship with the help of latent variables. Here, each latent variable is encoded to account for the various visual properties of the input data. The authors project the visual embedding to the semantic embedding space. 

\myparagraph{BiDiLEL \cite{bidilel}} projects both the visual and semantic embeddings into a common latent space (instead of projecting to the semantic space) so that the intrinsic relationship between them is maintained. 

\myparagraph{SYNC \cite{sync}} uses a weighted bipartite graph in order to learn a projection between the semantic embeddings and the classifier model space. They generate the graph by using a set of "phantom" classes synthesized in order to ensure aligned semantic embedding space and classifier model space and minimize the distortion error.

\myparagraph{WGAN \cite{wgan}} uses a Wasserstein GAN to synthesize the unseen features of classes, with additional losses in the form of cosine and cycle-consistency losses. These losses help enhancing the feature generation process. 

\myparagraph{OD \cite{od}} trains an out-of-distribution detector to distinguish the generated features from those of the seen class features and in turn to help with classification in the generalized zero-shot learning setting.

\myparagraph{E2E \cite{e2e}} is a recent approach that leverages end-to-end training to alleviate the problem of overlapping classes. This is done by removing all overlapping classes in the pre-training dataset and then using a CNN trained on the remaining classes to generate the visual features for the ZSL videos. 


\myparagraph{CLASTER \cite{claster}} uses clustering of visual-semantic embeddings optimised by reinforcement learning.

\section{Evaluation protocol}
\label{sec:eval}
\subsection{Datasets}

The three most popular benchmarks for ZSL in videos are UCF101 \cite{ucf101}, HMDB51 \cite{hmdb} and Olympics \cite{olympics}. 
The typical evaluation protocol in video ZSL is to use a 50-50 split of each dataset, where 50 \% of the labels are used as the training set and 50 \% as the test set. 
In order to provide comparisons to prior work \cite{bidilel,latem,sync,od,e2e,claster} and for the purpose of communicating replicable research results, we study UCF101, HMDB51, and Olympics, as well as the the relationship to the pre-training dataset Kinetics-400.

In our experiments (see Section~\ref{sec:exp}), we find overlapping classes between Kinetics-400 and each of the ZSL datasets, and move them to the training split. Thus, instead of using 50-50, we need to use 70-31 (number of labels for train and test) for UCF101 and 29-22 (number of labels for train and test) for HMDB51. We see that the number of overlapping classes in the case of Olympics is 13 out of 16, and hence we choose not to proceed further with it. More details can be found in Table~\ref{tab:datasetsplits}. For a fair comparison between the \newsplitname and random split, we use the same proportions (i.e., 70-31 in UCF101 and so on) in the experiments with random splits. 
We create ten such random splits and use these same splits for all models.


\subsection{\newsplitname Split}
\label{sec:truezeroshotsplit}

\begin{table}[t]
\begin{center}
\begin{tabular}{|l|c|c|c|c|c|}
\hline
Dataset & Videos & Classes & Random Split* & Overlapping classes & \newsplitname Split \\
& & & (Seen/Unseen) & with Kinetics & (Seen/Unseen) \\
\hline\hline
Olympics & 783 & 16 & 8/8 & 13 & - \\
HMDB51 & 6766 & 51 & 26/25 & 29 & 29/22\\
UCF101 & 13320 & 101 & 51/50 & 70 & 70/31\\
\hline
\end{tabular}
\end{center}
\caption{Datasets and their splits used for ZSL in action recognition. Traditionally, `Random Split' was followed where the seen and unseen classes were randomly selected. However, we can see the extent of overlap in the `overlapping classes' column. Using the extent of overlap we define our `\newsplitname split'.  For the full list of seen and unseen classes, please look at the supplementary. *Note that for the all experiments in this paper we use a random split which matches the number of classes of our \newsplitname, e.g. 29/22 for HMDB51.}
\label{tab:datasetsplits}
\end{table}



We now describe the process of creating the proposed \newsplitname split, to avoid the coincidental influence of pre-training on ZSL. 
First, we identify overlapping classes between the pre-training Kinetics-400 dataset and each ZSL dataset. To do this, we compute visual and semantic similarities, and discard those classes that are too similar. 

To calculate visual similarity, we use an I3D model pre-trained on Kinetics-400 and evaluate all video samples in UCF101, HMDB51 and Olympics using the Kinetics labels. This helps us to detect similarities that are often not recognized in terms of semantic similarities. Some examples include typing (class in UCF101) that the model detects as using computer (class in Kinetics), applying eye makeup (class in UCF101) that the model detects as filling eyebrows (class in Kinetics).

\begin{figure}[t]
    \centering
    \includegraphics[width=0.9\linewidth]{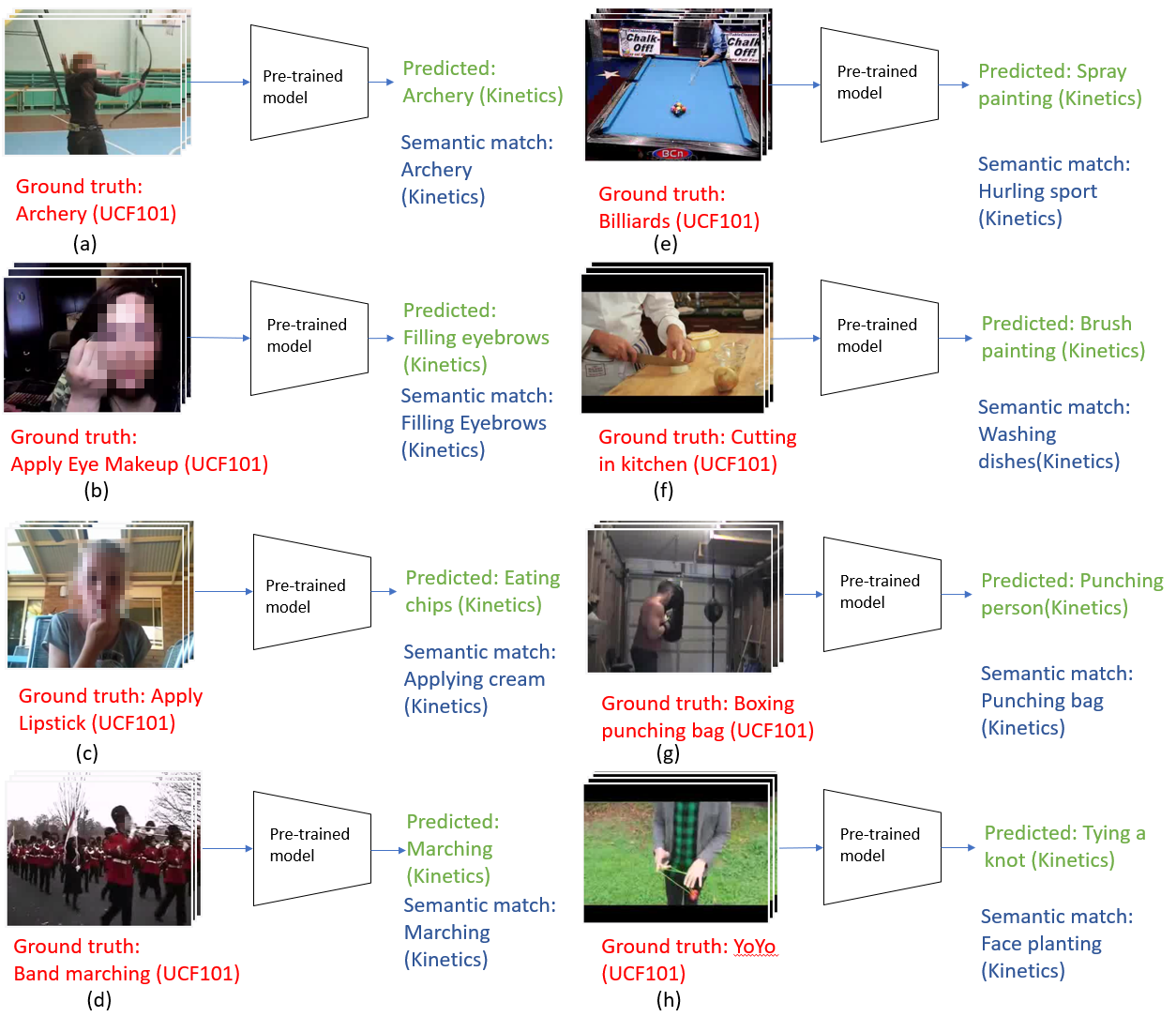}
    \caption{A few examples of how the classes are selected. (a) is an example of an exact match between the testing dataset (in this case UCF101) and the pre-trained dataset (Kinetics). (b), (d) and (g) are examples of visual-semantic similar matches where the output and semantically closest classes are the same. (c), (e), (f) and (h) are examples of classes without overlap in terms of both visual and semantic similarity.}
    \label{fig:vis_sem}
\end{figure}

To calculate semantic similarity, we use a sen2vec model pre-trained on Wikipedia that helps us compare action phrases and outputs a similarity score. We combine the visual and semantic similarity to obtain a list of extremely similar classes to the ones present in Kinetics. This list of classes is present in the supplementary. The classes that even have a slight overlap or are a subset of a class in Kinetics are all chosen as part of the seen set (for example, cricket bowling and cricket shot in UCF101 are part of the seen set due to the superclass playing cricket in Kinetics). A few examples of the selection of classes is show in Figure~\ref{fig:vis_sem}.

We discard classes from the test set based on the following rules:
\begin{itemize}
    \item Discard exact matches. For example, archery in UCF101 is also present in Kinetics.
    \item Discard matches that can be either superset or subset. For example, UCF101 has classes such as cricket shot and cricket bowling while Kinetics has playing cricket (superset). We manually do this based on the output of the closest semantic match.
    \item Discard matches that predict the same visual and semantic match. For example, \emph{apply eye makeup} (UCF101 label) predicts \emph{filling eyebrows} as the visual match using Kinetics labels and the closest semantic match to classes in Kinetics is also \emph{filling eyebrows}. We also manually confirm this.
\end{itemize}

We move all the discarded classes to the training set. This leaves a 70-31 split on UCF101 and a 29-22 split on HMDB51. We also see that in the Olympics dataset, there are 13 directly overlapping classes out of 16 classes and hence dropped the dataset from further analysis. One particular interesting scenario is the ``pizza tossing" class in UCF101. In Kinetics, there is a class called ``making pizza", however, the action of tossing is not performed in them and hence we use ``pizza tossing" as an unseen class.

\section{Experimental Results}
\label{sec:exp}

\subsection{Results on ZSL and Generalized ZSL}

We first consider the results on ZSL. Here, as explained before, only samples from the unseen class are passed as input to the model at test time. Since \newsplitname separates the overlapping classes from the pre-training dataset, we expect a lower accuracy on this split compared to the traditionally used random splits. We compare BiDiLEL \cite{bidilel}, Latem \cite{latem}, SYNC \cite{sync}, OD \cite{od}, E2E \cite{e2e} and CLASTER \cite{claster} and report the results in Table~\ref{tab:zsl}. As expected, we see in the `Diff' column for both UCF101 and HMDB51 a positive difference, indicating that the accuracy is lower for the \newsplitname split.

\begin{table}[t]
\begin{center}
\begin{tabular}{|l|ccc|ccc|}
\hline
Method & \multicolumn{3}{|c|}{UCF101}  &  \multicolumn{3}{|c|}{HMDB51}  \\
 & Random & \newsplitname & Diff & Random & \newsplitname & Diff \\
\hline\hline

Latem \cite{latem} & 21.4 & 15.5 & 5.9 & 17.8 & 9.4 & 8.4\\
SYNC \cite{sync} & 22.1 & 15.3 & 6.8 & 18.1 & 11.6 & 6.5\\
BiDiLEL \cite{bidilel} & 21.3 & 15.7 & 5.6 & 18.4 & 10.5 & 7.9\\
OD \cite{od} & 28.4 & 22.9 & 5.5 & 30.6 & 21.7 & 8.9\\
E2E \cite{e2e} & 46.6 & 45.5 & 1.1 & 33.2 & 31.5 & 1.7\\
CLASTER \cite{claster} & 47.1 & 45.2 & 1.9 & 36.6 & 33.2 & 3.4\\
\hline
\end{tabular}
\end{center}
\caption{Results with different splits for \textbf{Zero-Shot Learning (ZSL)}. Column `Random' corresponds to the accuracy using splits in the traditional fashion (random selection of train and test classes, but with the same number of classes in train/test as in \newsplitname), `TruZe' corresponds to the accuracy using our proposed split and `Diff' corresponds to the difference in accuracy between using random splits and our proposed split. We run 10 independent runs for different random splits and report the average accuracy. We see positive differences in the `Diff' column which we believe is due to the overlapping classes in Kinetics. }
\label{tab:zsl}
\end{table}

\begin{figure}[t]
    \centering
    \includegraphics[width=0.95\linewidth]{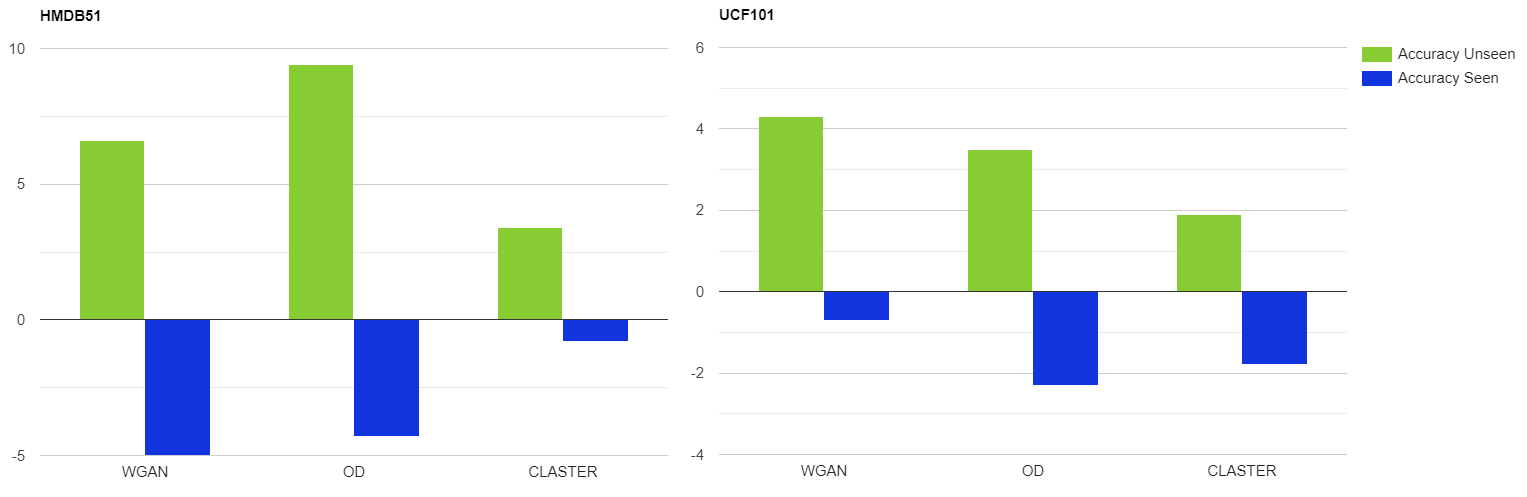}
    \caption{Graphical representation of the difference in performances of different models on GZSL. We see consistent positive difference in performance on the unseen classes and negative difference in the performance of the seen classes while using the `\newsplitname'. The x-axis corresponds to difference in accuracy (Random splits accuracy - \newsplitname split accuracy) and the y-axis to different methods.}
    \label{fig:bar:hmdb:gzsl}
\end{figure}

Generalized ZSL (GZSL) looks at a more realistic scenario, wherein the samples at test time belong to both seen and unseen classes. The reported accuracy is then the harmonic mean of the seen and unseen class accuracies. Since we separate out the overlapping classes, we expect to see an increase in the seen class accuracy and a decrease in the unseen class accuracy. We report GZSL results on OD, WGAN and CLASTER in Table~\ref{tab:gzsl}. The semantic embedding used for all models is sen2vec. We use 70 classes for training chosen at random along with 31 test classes (also chosen at random) for UCF101 and 29 training with 22 testing for HMDB51. As expected, the average unseen class accuracy drops in the proposed split and the average seen class accuracy increases. We expect this as the unseen classes are more disjoint in the proposed split than using random splits. For easier understanding, we convert the differences in Table~\ref{tab:gzsl} to a graph and this can be seen in Figure~\ref{fig:bar:hmdb:gzsl}.

\begin{table}[t]
\begin{center}
\begin{tabular}{|l|ccc|ccc|ccc|c|}
\hline
 &  \multicolumn{3}{|c|} {Acc$_{U}$}  & \multicolumn{3}{|c|} {Acc$_{S}$}  & \multicolumn{3}{|c|}{Harmonic mean} & \\
Method & Rand & \newsplitname & Diff & Rand & \newsplitname & Diff & Rand & \newsplitname & Diff &  Dataset\\
\hline\hline
WGAN \cite{wgan} &  27.9 & 21.3 & 6.6 & 58.2 & 63.2 & -5.0 & 37.7 & 31.8 & 5.9 & HMDB51\\
WGAN \cite{wgan} &  28.2 & 23.9 & 4.3 & 74.9 & 75.6 & -0.7 & 41.0 & 36.3 & 4.7 & UCF101\\
\hline
OD \cite{od} & 34.1 & 24.7 & 9.4 & 58.5 & 62.8 & -4.3 & 43.1 & 35.5 & 7.6 & HMDB51\\
OD \cite{od} & 32.6 & 29.1 & 3.5 & 76.1 & 78.4 & -2.3 & 45.6 & 42.4 & 3.2 & UCF101\\
\hline
CLASTER \cite{claster} & 41.8 & 38.4 & 3.4 & 52.3 & 53.1 & -0.8 & 46.4 & 44.5 & 1.9 & HMDB51\\
CLASTER \cite{claster} & 37.5 & 35.6 & 1.9 & 68.8 & 70.6 & -1.8 & 48.5& 47.3 & 1.2 & UCF101\\
\hline
\end{tabular}
\end{center}
\caption{Results with different splits for \textbf{Generalized Zero-Shot Learning (GZSL)}. `Rand' corresponds to the splits using random classes over 10 independent runs, `\newsplitname' corresponds to the proposed split. Acc$_{U}$ and Acc$_{S}$ correspond to unseen class accuracy and seen class accuracy respectively. The semantic embedding used is sen2vec. `diff' corresponds to the difference between `Rand' and `\newsplitname'. We see consistent positive difference in performance on the unseen classes and negative difference in the performance of the seen classes while using the `\newsplitname'.}
\label{tab:gzsl}
\end{table}

\subsection{Extension to Few-shot Learning}

\begin{table}
\begin{center}
\begin{tabular}{|l|ccccc|ccccc|ccccc|}
\hline
Method &  \multicolumn{5}{|c|} {SS}  & \multicolumn{5}{|c|} {\newsplitname}  & \multicolumn{5}{|c|} {Diff} \\
& 1 & 2 & 3 & 4 & 5 & 1 & 2 & 3 & 4 & 5 & 1 & 2 & 3 & 4 & 5 \\
\hline\hline

C3D-PN \cite{pn} & 57.1 & 66.4 & 71.7 & 75.5 & 78.2 & 50.9 & 61.9 & 67.5 & 72.9 & 75.4 & 6.2 & 4.5 & 4.2 & 2.6 & 2.8\\
\hline
ARN \cite{arn} & 66.3 & 73.1 & 77.9 & 80.4 & 83.1 & 61.2 & 70.7 & 75.2 & 78.8 & 80.2 & 5.1 & 2.4 & 2.7 & 1.6 & 2.9\\
\hline
TRX \cite{perrett2021trx} & 77.5 & 88.8 & 92.8 & 94.7 & 96.1 & 75.2 & 88.1 & 91.5 & 93.1 & 93.5 & 2.5 & 0.7 & 1.3 & 1.6 &2.6 \\

\hline
\end{tabular}
\end{center}
\caption{\textbf{Few Shot Learning (FSL)} with different splits on UCF101. Accuracies are reported for 5-way, 1, 2, 3, 4, 5-shot classification. 'SS' corresponds to the split used in \cite{arn,perrett2021trx} and '\newsplitname' corresponds to the proposed split. We can see that using our proposed split results in a drop in performance of up to 6.2 \% for UCF101. This shows \newsplitname is much harder even in the FSL scenario.}
\label{tab:fslucf}
\end{table}

\begin{table}
\begin{center}
\resizebox{\columnwidth}{!}{%
\begin{tabular}{|l|ccccc|ccccc|ccccc|}
\hline
Method &  \multicolumn{5}{|c|} {SS}  & \multicolumn{5}{|c|} {\newsplitname}  & \multicolumn{5}{|c|} {Diff} \\
& 1 & 2 & 3 & 4 & 5 & 1 & 2 & 3 & 4 & 5 & 1 & 2 & 3 & 4 & 5 \\
\hline\hline

C3D-PN \cite{pn} & 38.1 & 47.5 & 50.3 & 55.6 & 57.4 & 28.8 & 38.5 & 43.4 & 46.7 & 49.1 & 9.3 & 9.0 & 6.9 & 8.9 & 8.3\\
\hline
ARN \cite{arn} & 45.5 & 50.1 & 54.2 & 58.7 & 60.6 & 31.9 & 42.3 & 46.5 & 49.8 & 53.2 & 12.6 & 7.8 & 7.7 & 8.9 & 7.4\\
\hline
TRX \cite{perrett2021trx} & 50.5 & 62.7 & 66.9 & 73.5 & 75.6 & 33.5 & 46.7 & 49.8 & 57.9 & 61.5 & 17.0 & 16.0 & 17.1 & 15.6 & 14.1\\

\hline
\end{tabular}%
}
\end{center}
\caption{\textbf{Few Shot Learning (FSL)} with different splits on HMDB51. Accuracies are reported for 5-way, 1, 2, 3, 4, 5-shot classification. 'SS' corresponds to the split used in \cite{arn,perrett2021trx} and '\newsplitname' corresponds to the proposed split. We can see that using our proposed split results in a drop in performance of up to 17.1 \% for HMDB51. This shows \newsplitname is much harder even in the FSL scenario.}
\label{tab:fslhmdb}
\end{table}

Few-shot learning (FSL) is another scenario we consider. Since the premise is the same as ZSL, except that we have a few samples instead of zero. Again, usually, the splits used are random, and as such, the pre-trained model has seen hundreds of samples of classes that are supposed to belong to the test set. We report results on the 5-way, 1,2,3,4,5-shot case for temporal relational cross-transformers (TRX) \cite{perrett2021trx}, action relation network (ARN) \cite{arn}, and C3D prototypical net (C3D-PN) \cite{pn}. Results are reported in Table~\ref{tab:fslucf} and Table~\ref{tab:fslhmdb}. The standard split (SS) used here is taken from the one proposed in ARN \cite{arn}. Similar to the SS, we divide the classes in UCF101 and HMDB51 to (70,10,21) and (31,10,10), respectively, where the order corresponds to the number of training classes, validation classes and test classes. We see that the proposed split is much harder than SS. Consistent drops in performance can be seen on every split and for every model. Performance drops of upto 6.2\% on UCF101 and 17.1\% on HMDB51 can be seen. Our proposed splits are available in the supplementary material. 

\subsection{Is overlap the reason for performance difference between Random and our \newsplitname split?}

\begin{figure}[t]
    \centering
    \includegraphics[width=0.95\linewidth]{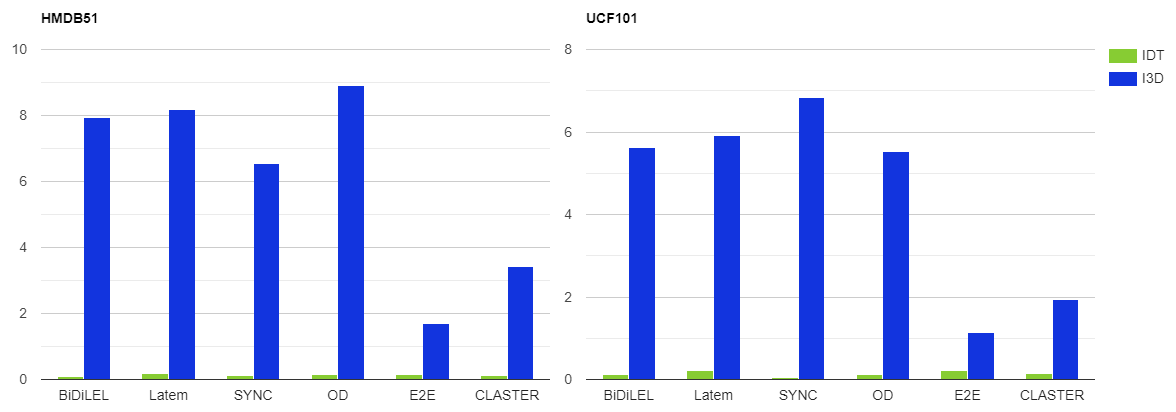}
    \caption{The difference of accuracy for different models using IDT and I3D using manual annotations as the semantic embedding. The larger the bar, the more significant the difference. We can see a clear difference when using I3D and this difference is due to the presence of overlapping classes in the test set. The y-axis corresponds to the difference in performance in percentage and the x-axis corresponds to various models.}
    \label{fig:bar}
\end{figure}

In order to understand the difference in model performance due to the overlapping classes, we compare the performance of each model for the random split (with five runs) vs the proposed split by using visual features represented by IDT and I3D. We depict the difference in performance in the form of a bar graph for better visual understanding. This is seen in Figure~\ref{fig:bar}. The higher the difference, the bigger the impact of performance. We can see that there is a big difference when using I3D features compared to using IDT features (where there is a minimal difference). Since IDT features are independent of any pre-training model, the difference in performance is negligible. The difference while using I3D features can be attributed to the presence of overlapping classes in the random splits compared to the proposed split.

\subsection{Use of different backbone networks}
\label{sec:backbone}

An end-to-end approach was proposed in \cite{e2e} where a 3D CNN was trained in an end-to-end manner on classes in Kinetics not semantically similar to UCF101 and HMDB51 to overcome the overlapping classes conundrum. While this approach is useful, training more complex models end-to-end is not feasible for everyone due to the high computational cost involved. We show that using more recent state-of-the-art approaches as the backbone, there is a slight improvement in model performance and hence believe that having a proposed split instead of training end-to-end would be more easily affordable for the general public. Table~\ref{tab:backbone} shows the results of using different backbones for extracting visual features on some of the recent state-of-the-art ZSL approaches. We use Non-Local networks \cite{nl} that build on I3D by adding long-term spatio-temporal dependencies in the form of non-local connections (referred as NL-I3D in Table~\ref{tab:backbone}). We also use slow-fast networks \cite{slowfast} that is a recent state-of-the-art approach that uses two pathways, a slow and a fast, to capture motion and fine temporal information. We can see minor but consistent improvements using stronger backbones, and this suggests that having a proposed split is an economical way of maximising the use of state-of-the-art models as backbone networks. We see gains of up to 0.6\% in UCF101 and 0.8\%  in HMDB51.

\begin{table}[t]
\begin{center}
\begin{tabular}{|l|c|c|c|}
\hline
Method & Backbone & UCF101 Accuracy & HMDB51 Accuracy \\
\hline\hline
WGAN \cite{wgan} & I3D & 22.5 & 21.1 \\
WGAN \cite{wgan} & NL-I3D & 22.7 & 21.3 \\
WGAN \cite{wgan} & SlowFast & \textbf{23.1} & \textbf{21.5} \\
\hline
OD \cite{od} & I3D & 22.9 & 21.7 \\
OD \cite{od} & NL-I3D & 23.2 & 22.0 \\
OD \cite{od} & SlowFast & \textbf{23.4} & \textbf{22.5} \\
\hline
CLASTER \cite{claster} & I3D & 45.2 & 33.2 \\
CLASTER \cite{claster} & NL-I3D & 45.3 & 33.6 \\
CLASTER \cite{claster} & SlowFast & \textbf{45.5} & \textbf{33.9} \\
\hline
\end{tabular}
\end{center}
\caption{Results comparison using different backbones to extract visual features for the ZSL models. We evaluate OD, E2E and CLASTER using I3D, NL-I3D and SlowFast networks as backbones. All results are on the proposed split. We see that stronger backbones result in improved performance of the ZSL model.}
\label{tab:backbone}
\end{table}

\section{Implementation Details}

\subsection{Visual features}

We use either IDT \cite{idt} or I3D \cite{i3d} for the visual features. Using the fisher vector obtained from a 256 component Gaussian mixture model, we generate visual feature representations using IDT (contains four different descriptors). To reduce this, PCA is used to obtain a 3000-dimensional vector for each descriptor. Concatenating these (all four descriptors), we obtain a 12000-dimensional vector for each video. In the case of I3D features, we use RGB and flow features taken from the $\textit{mixed 5c}$ layer from a pre-trained I3D (pre-trained on Kinetics-400). 
The output of the flow network is averaged across the temporal dimension and
pooled by four in the spatial dimension, and then flattened to
a vector of size 4096. We then concatenate the two.

\subsection{Semantic embedding}

While manual annotations are available for UCF101 in the form of a vector of size 40, there is no such annotation available for HMDB51. Hence, we use sen2vec embeddings of the action classes where the sen2vec model is pre-trained on Wikipedia. While most approaches use word2vec and average embeddings for each word in the label, we use sen2vec which obtains an embedding for the entire label.

\subsection{Hyperparameters for evaluated methods}

We use the optimal parameters reported in BiDiLEL \cite{bidilel}. The values for $\alpha$ and k$_{G}$values are set to 10 and d$_{y}$ is set to 150. SYNC \cite{sync} has a parameter $\sigma$ that models correlation between a real class and a phantom class and this is set to 1, while the balance coefficient is set to 2$^{-10}$. For Latem \cite{latem} the learning rate,  number of epochs,  and number of embeddings are 0.1, 200 and 10 respectively. For OD \cite{od}, WGAN \cite{wgan}, E2E \cite{e2e} and CLASTER \cite{claster} we follow the settings provided by the authors. 
For few-shot learning, we use the hyperparameters defined in the papers \cite{perrett2021trx,arn}. We compare against the standard split proposed in \cite{arn}. For the proposed split, we change the classes slightly for fair comparison to the standard split. Now the splits for HMDB51 and UCF101 are (31,10,10) and (70,10,21) where the order corresponds to (train,val,test).

\section{Discussion and Conclusion}


As we see in Figure~\ref{fig:bar} using IDT features which do not require pre-training on Kinetics resulted in a negligible change in performance comparing the \newsplitname split vs the random splits. However, using I3D features saw a stark difference due to the overlapping classes in the pre-trained dataset.



We see that the proposed split is harder in all scenarios (ZSL, GZSL, and FSL) whilst maintaining the premise of the problem. The differences are significant in most cases: between 0.7-6.2 \% for UCF101 and 7.4-17.1 \% for HMDB51 in FSL, an increase of 1.2-4.7 \% for UCF101 and 1.9-7.6 \% for HMDB51 (with respect to the harmonic mean of seen and unseen classes) in GZSL and an increase of 1.1-6.8 \% for UCF101 and 1.7-8.1 \% for HMDB51 in ZSL. It is also important to note that different methods are differently affected, suggesting that some method in the past have claimed improvements due to not adhering to the zero-shot premise, which is highly concerning.

We also see that changing the backbone network increases the performance slightly for each model, and as a result, the end-to-end pre-training \cite{e2e} can prove very expensive. As such, having a proposed split makes things easier as we can directly use pre-trained models off the shelf. We see gains of up to 1.3\% in UCF101 and 1.1\%  in HMDB51.

Details to our \newsplitname splits can be found in supplemental material and we will release them publicly so the research community can fairly compare zero-shot and few-shot action recognition approaches and compare to the benchmark results provided in this paper.
\bibliographystyle{splncs04.bst}
\bibliography{069-main}

\end{document}